\documentclass[journal]{IEEEtran}
\ifCLASSINFOpdf
  \usepackage[pdftex]{graphicx}
  
\else
\fi
\usepackage{amsmath}
\usepackage{amssymb}
\usepackage{booktabs}
\usepackage{multirow}
\begin{document}
%
\title{Large Margin Mechanism and Pseudo Query Set\\ on Cross-Domain Few-Shot Learning}
%
%
%

\author{Jia-Fong Yeh,~\IEEEmembership{Student Member,~IEEE,} 
        Hsin-Ying Lee, 
        Bing-Chen Tsai*,
        Yi-Rong Chen*,\\
        Ping-Chia Huang*, 
        Hung-Ting Su,~\IEEEmembership{Student Member,~IEEE,}
        and~Winston H. Hsu,~\IEEEmembership{Senior Member,~IEEE}%
        \thanks{All authors are affiliated with National Taiwan University, and the authors marked with * contributed equally to this work.}%
        \thanks{Email: jiafongyeh@ieee.org, \{b05705022, b06902066, b06902001, b06902024\}@ntu.edu.tw, htsu@cmlab.csie.ntu.edu.tw, whsu@ntu.edu.tw}
        }
        
%
%

%
%

\markboth{IEEE TRANSACTIONS ON MULTIMEDIA,~Vol.~XX, No.~X, MONTH~YEAR}%
{Yeh \MakeLowercase{\textit{et al.}}: IEEE TRANSACTIONS ON MULTIMEDIA}
%



\maketitle

\begin{abstract}
   In recent years, \textit{few-shot learning (FSL)} have received a lot of
   attention due to difficulties and costs of data collection for many real-world problems.
   Conventional approaches focus on single-domain FSL where the base classes and novel
   classes are mostly from the same domain. We study a more challenging yet practical task,
   \textit{cross-domain few-shot learning (CD-FSL)}, which aims to recognize classes that
   are not only unseen but also from a different domain. We propose a novel large margin fine-tuning
   method (LMM-PQS), which enables models pre-trained on a single domain to be adaptable to various
   different domains with only a few labeled target samples. Our LMM-PQS centers on three novel
   components. (1) Pseudo query set (PQS): we generate more data to solve the problem that
   the query set is not available during fine-tuning. (2) Prototypical triplet loss (PT loss):
   it is a modified triplet loss integrating with the prototype concept which is commonly used in few-shot 
   models. (3) Large margin cosine loss (LMCL): we treat CD-FSL problem as a 
   \emph{near-open-set problem}, and thus the large margin cosine loss is applied. Our LMM-PQS significantly
   outperforms competitive baselines on various different target domains, which demonstrates promising 
   effectiveness on challenging CD-FSL problem. Most importantly, we provide a practical solution for CD-FSL with only a few data available. 
\end{abstract}


%
\IEEEpeerreviewmaketitle

\section{Introduction}
%
%
%
%
\IEEEPARstart{G}{eneral} deep neural networks (DNNs) \cite{Krizhevsky12AN} for computer vision and multimedia
problems heavily rely on a large amount of labeled training data and need to be re-trained or
fine-tuned when encountering different tasks. Besides, the generalization ability of these
networks is highly correlated with the diversity and the size of the training set. However, collecting
adequate amounts of data for practical problems is usually difficult and high-cost. Therefore, learning to characterize different classes with a few labeled samples, known as 
\emph{few-shot learning}, is necessary. The problem is usually set under the meta-learning setting,
which contains meta-training and meta-testing phases. In the first phase, models are trained
with classification tasks where data composed of labeled samples, known as support sets, and 
unlabeled samples, known as query sets, both sampled from \emph{base} classes. In the second
phase, tasks sampled from \emph{novel} classes are used for evaluating the model performance. Note that \emph{base} classes and
\emph{novel} classes are disjoint, and thus quick adaptation from \emph{base} classes to 
\emph{novel} classes is indispensable and challenging. 

The objective of FSL problem is to categorize the query set leveraging  given support set in the meta-testing phase. Myriads of few-shot learning approaches \cite{Vinyals16MN, Snell17PN, Sung18RN, Zhang19VSL, Wu19PARN} try designing different
metrics to recognize the relationship between two sets. However, a recent work \cite{Chen19Baseline} claims that
a general DNN, denoted as Baseline, trained under a standard practice shows
competitive results with several fine-tuning iterations. 

In the training procedure, the biggest difference between general DNNs (e.g.\ Baseline) and few-shot methods is that the former 
trains and updates parameters with batch data at every iteration, whereas the latter makes 
inference based on support sets (a few labeled data) and updates parameters with loss over query sets.
As mentioned above, general DNNs need to be fine-tuned before solving novel tasks. Moreover, the query set is
reserved for evaluation in the meta-testing phase, which means only support set can be used for fine-tuning.
Consequently, general DNNs can fine-tune using data sliced from support set and adapt to novel tasks.
On the other hand, few-shot methods infer the query set directly without fine-tuning. The reasons
why few-shot methods don't apply fine-tuning are (1) both support set and query set are needed for updating their parameters, and (2) they are trained to learn the relationship between support set and query set rather than how to classify query set, therefore, they should be flexible to adapt to novel tasks. From the literature, general DNNs show similar results with common few-shot models under conventional few-shot learning problem \cite{Chen19Baseline}.

\begin{figure*}[!t]
  \centering
  \includegraphics[width=\textwidth]{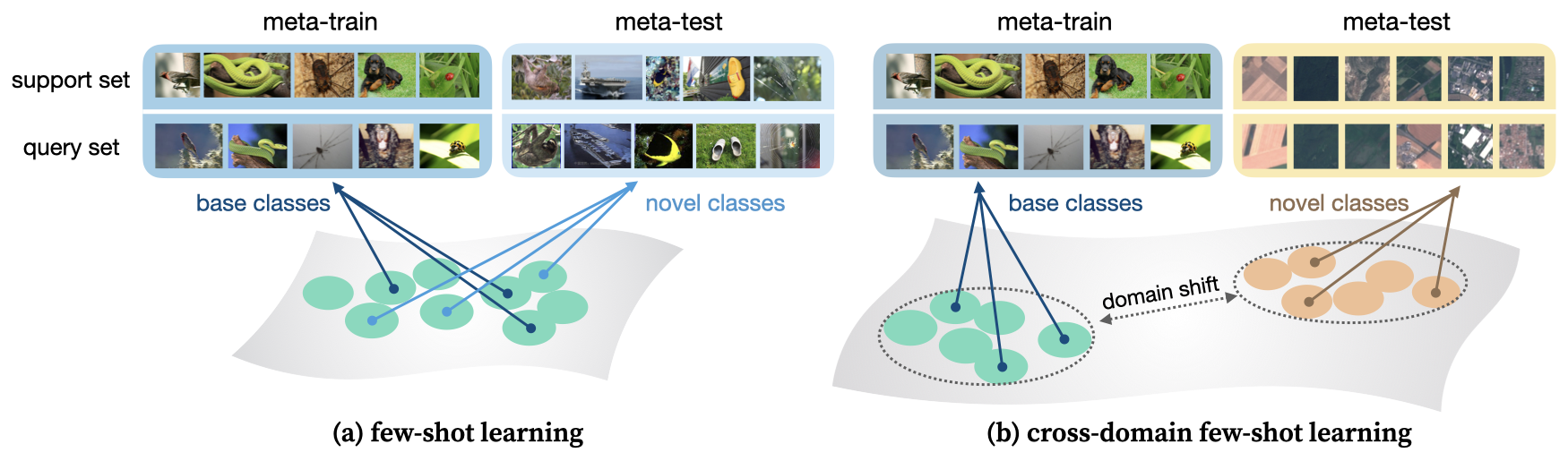}
  \caption{Illustration of general and cross-domain few-shot learning problem. Few-shot learning problem is usually under
  meta-learning scenario, which includes meta-training and meta-testing phase. Models face a new task in every
  iteration. A task contains support set (few labeled data) and query set (unlabeled
  data). The goal of few-shot learning aims to infer the query set label with the guidance from support set. (a) In conventional \textit{few-shot learning}, tasks in the meta-training phase are sampled from \textit{base} classes and task in the meta-testing phase are sampled from \textit{novel} classes. The two classes are disjoint, but they might be similar to each other (e.g. two subsets from the same dataset). (b) In \textit{cross-domain few-shot learning}, 
  \textit{base} classes and \textit{novel} classes could be in an absolutely different domain. For example, \textit{base}
  classes are sampled from common object images, and \textit{novel} classes are sampled from satellite images. (cf. Section \ref{section_problem_definition})}
  \label{fig:teaser}
\end{figure*}

Moreover, some few-shot models also conduct experiments under the cross-domain setting where a 
domain shift exists between \emph{base} classes and \emph{novel} classes (cf. Figure 
\ref{fig:teaser}(b)), such as meta-training on mini-ImageNet \cite{Vinyals16MN} and meta-testing 
on CUB dataset \cite{Hilliard18CUB}. However, compared to human's ability and daily experience of 
dealing with problems dissimilar from ones' knowledge, these two datasets, as both contains common
objects, only demonstrate machines' capability in a limited scenario. Even more interesting is few
shot methods without further fine-tuning show inferior performance compared to general DNNs under cross-domain setting in \cite{Guo19ANB}.

Therefore, in this work, we further investigate the \emph{cross-domain few-shot learning problem (CD-FSL)}, especially when huge difference lies between domains of \emph{base} classes and \emph{novel} classes. We hypothesize that few-shot methods still need to fine-tune when a huge domain shift exited, and thus we propose a fine-tuning method for few-shot models which generates pseudo query images as an alternative and \textit{without access to the query set} during fine-tuning. With the pseudo query set, few-shot models can fine-tune using the same style
as in the meta-training phase. The parameter updating is consistence in both phases, which is the most important contribution of our work. Moreover, we handle the CD-FSL problem
as a \emph{near-open-set problem} \footnote{The CD-FSL problem is not a open-set problem  because the labeled support set is given in the meta-testing phase.}, and large margin mechanism is widely used in open-set 
problems, so we leverage two large margin mechanisms to improve the performance, consisting of a 
novel prototypical triplet loss and large margin cosine loss (motivated by open-set face recognition models \cite{Wang18CosFace}). It is worth noting that we deal with CD-FSL problems under agnostic setting, which
means that the information of query set (including features and label) is not used during fine-tuning. We need
to infer the query set immediately when first accessing them. 

\begin{figure*}[t]
  \centering
  \includegraphics[width=\textwidth]{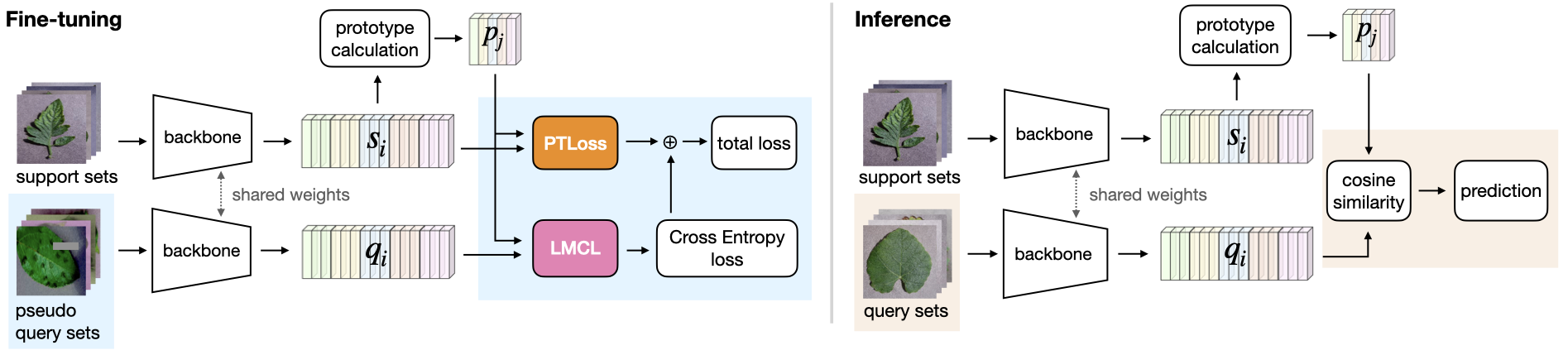}
  \caption{Overview of large margin mechanism and pseudo query set (LMM-PQS). Meta-testing phase consists of two stages, fine-tuning and 
  inference. At fine-tuning stage, LMM-PQS first generates pseudo query set (PQS, cf. Section \ref{section_pqs}) and uses trained backbone to
  compute the feature embeddings. After calculating the class prototypes, LMM-PQS applies PT loss (cf. Section \ref{PT 
  loss}) to enlarge the inter-class distance and decrease the intra-class distance. Moreover, LMM-PQS also executes LMCL
  (a large margin mechanism, cf. Section \ref{section_lmcl}) to enhance the inferring confidence. At inference stage,
  the progress before prototype calculation is the same as at fine-tuning stage. Then a cosine mean-centroid classifier
  compares the cosine similarity between query sample and class prototypes to categorize the class. See Section
  \ref{methodology} and Figure \ref{fig_pqs}-\ref{fig_lmm} for detailed description.}
  \label{flow_LMM-PQS}
\end{figure*}

Our main contributions are fourfold:
\begin{itemize}
\item \textbf{Pseudo query set.} To get out of the predicament that few-shot methods can't fine-tune, we propose the pseudo query set and analyze its importance.
\item \textbf{Prototypical triplet loss.} In order to meake the model more suitable in the FSL problem, we modify the triplet loss and propose a novel prototypical triplet loss.
\item \textbf{Large margin cosine loss.} Inspired by the similarity between open-set and cross-domain few-shot learning problem, the large margin cosine loss is applied.
\item \textbf{Detailed comparison and visualization results.} We conduct a comprehensive comparison between general DNNs and few-shot models on CD-FSL problem, and provide several visualization results to elaborate the robustness of proposed LMM-PQS method.
\end{itemize}

The rest of this paper is organized as follows. Section II reviews  studies from multiple related research area. Section III elaborates the CD-FSL problem and proposed LMM-PQS method. Experiment results are illustrated in Section IV. Moreover, we provide a further discussion in Section V, and the conclusion is made in Section VI. 

\section{Related Work}
\subsection{Few-shot classification} 
Few-shot classification is a task to recognize \emph{novel} classes with only a few 
labeled examples, and usually formulated as a meta-learning problem. Before solving
the \emph{novel} tasks, methods can learn knowledge from \emph{base} classes. There are 
two main factions of meta-learning approaches for addressing the few-shot classification
problem, including optimization based and metric-learning based methods.
\paragraph{Optimization based methods.}
Optimization based methods aims to learn a good initial configuration (a set of neural
parameters) for fine-tuning on few-shot problems. Given a network architecture, MAML
\cite{Finn17MAML} meta-learns the network initialization parameters and fine-tunes the network
on current task with single fine-tuning iteration. Then, the gradient is
calculated to update the initial parameters. Besides, Reptile \cite{Nichol18Reptile} is a first-order approximation of MAML. By ignoring second-order derivatives, Reptile has a faster execution speed.
\paragraph{Metric-learning based methods.}
Metric-learning based methods try to learn a general feature space and use various metric
to categorize unlabeled samples. MatchingNet \cite{Vinyals16MN} applies a 
bi-directional recurrent network and a weighted nearest neighbor classifier to recognize
the samples. ProtoNet \cite{Snell17PN} selects euclidean distance as the evaluation metric.
It calculates the class prototypes by the mean of the embedding of labeled samples and compares
the distance between unlabeled samples with prototypes. In addition, RelationNet \cite{Sung18RN}
is based on the same concept and introduces a learnable similarity metric. Many few-shot
approaches belong to this faction due to its simplicity and effectiveness.

Conventional few-shot image classification is well studied. In fact, many few-shot
approaches are designed to solve not only classification problems but also various real-world
applications \cite{Zou20AO, Huang20LR, Cao19MS}. We study a more challenging task, the cross-domain
few-shot learning, which is still scarcely investigated but urgently needed by many practical applications.

\subsection{Cross-domain few-shot learning}
Cross-domain few-shot learning is a new branch of few-shot learning, where \emph{base}
and \emph{novel} classes are sampled from different domains. In nowadays, the cross-domain
few-shot learning problem has not been well discussed yet, but we consider that it deserves
a further investigation due to its connection to real world problems. Compared to general
few-shot learning problem, Cross-domain few-shot learning is more challenging and laborious.
Also, we can make use of these data of different domains to measure the robustness and
generalization of few-shot approaches.

Tseng \textit{et al.} \cite{Tseng20Cross-Domain} propose a feature-wise transformation layer to
generalize image features for simulating different domains. By applying Gaussian distribution
after each BatchNorm layer, they try to mimic the features coming from various domains. Moreover,
the parameters $\gamma$ and $\beta$ of the Gaussian distribution are trained with pseudo tasks
generated by same or other datasets, depending on the problem setting. Experiment results show
that few-shot models with the feature-wise transformation layer can achieve a higher performance.
Besides, Guo \textit{et al.} \cite{Guo19ANB} propose a new benchmark for this problem. In this
benchmark, methods are trained on single domain dataset and meta-tested on various domain datasets. Moreover, a detailed comparison between several classifiers is provided. 

In order to overcome the domain shift between the base class and the novel class in CD-FSL problem, we propose the pseudo query set to let few-shot models can be fine-tuned. Moreover, compared to previous works, we treat CD-FSL problem as a near-open set problem. , and the large margin mechanism is popular in the open-set problem. Consequently, the novel PT loss and large margin cosine loss are applied to improve the performance.

\subsection{Large margin mechanism} \label{related_lmm}
Large margin mechanism is popular in the open-set problem. These approaches
aim to recognize unseen classes with sufficient margin during testing.
How to set the margin is the focus of these series of researches. Many models \cite{Schroff15Triplet, Liu17Sph, Wang18CosFace, Deng19ArcFace}  apply this mechanism
to learn highly discriminative features by maximizing the inter-class margin. 

In the early years, triplet loss \cite{Schroff15Triplet} made a huge success. Many approaches from
various research fields utilize triplet loss as one of the loss functions, also including the 
method in few-shot learning problem \cite{Wang18LargeMF}. Recently, CosFace \cite{Wang18CosFace} and 
ArcFace \cite{Deng19ArcFace} are two representative models. These two models attempt to classify
classes by the cosine similarity between normalized feature vector and normalized weight vector.
Although they add the margin according to different considerations, both their goals are to make the model
effectively distinguish between different categories.

We consider that \emph{cross-domain few-shot learning problem} has similar property with the 
open-set problem. Both of their approaches need to deal with the \textit{unseen}
samples in the testing phase. The similarity between these two problems is described in Section
\ref{section_lmcl}. Intuitively, CD-FSL problem is more difficult rather than traditional open-set problems.
For open-set problems, although the classes in the test phase are unseen during training, they are still similar classes (classes are from adjacent domains). In CD-FSL, the \textit{novel} classes might be extremely dissimilar with \textit{base}
classes. We are interested that if large margin mechanisms can assist few-shot models to solve
CD-FSL problems, and thus we take two mechanisms in our proposed LMM-PQS and investigate the performance.  

\section{Methodology} \label{methodology}
\subsection{Cross-Domain Few-Shot Learning Problem} \label{section_problem_definition}

In general few-shot scenario, a model \(f_{\theta} : \mathcal{X}\xrightarrow{}
\mathcal{Y}\) with parameters \(\theta\) is meta-trained on the tasks sampled from
\emph{base} classes data and meta-tested on the tasks from \emph{novel} classes
data. In practice, \emph{base} classes and \emph{novel} classes usually are two subsets from a large datasets. Furthermore. a task contains a support set \(S=\{x_{i}, y_{i}\}^{N \times K}_{i=1}\) and
a query set \(Q=\{x_{i}, y_{i}\}^{N \times M}_{i=1}\). This is known as an 
"\emph{N-way K-shot}" few-shot learning problem, as the support set has \emph{N}
classes and each class contains \emph{K} labeled samples. In addition, the query
set is used to evaluate the models inference performance, but the inference results
can be used to help model training during the meta-training phase.

\begin{figure}[t]
    \centering
    \includegraphics[width=.801\linewidth]{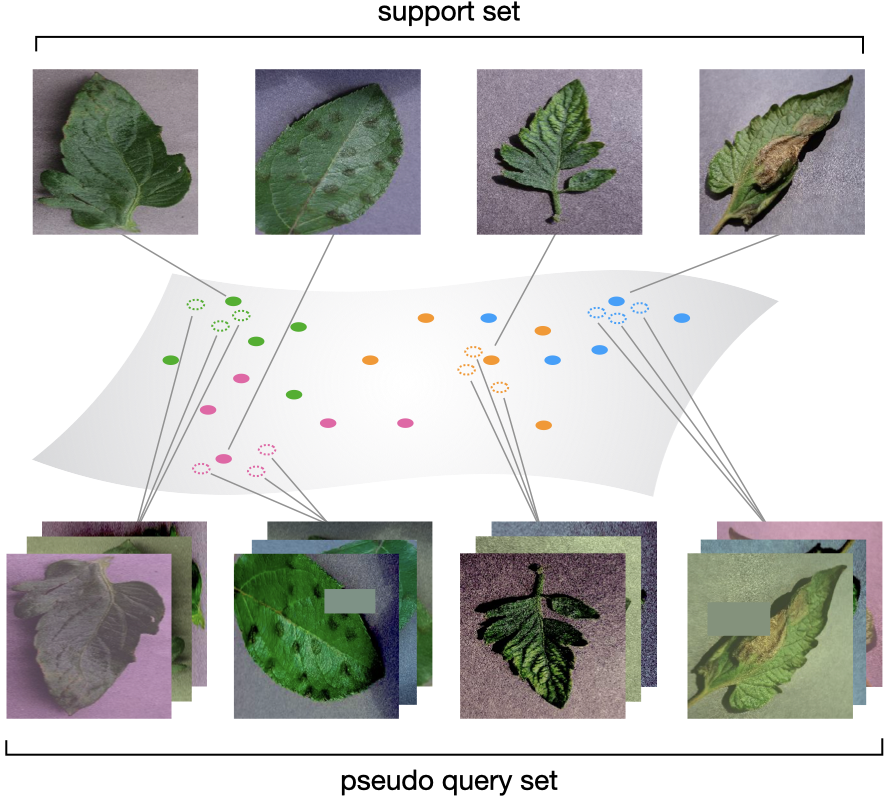}
    \caption{Pseudo query set (PQS). We use support samples (solid circles) to generate the pseudo query samples
    (hollow circles) by several digital image processing operations (cf. Section \ref{section_pqs}). With PQS,
    we can discover more feature space with a few target data and learn more knowledge. Most importantly, few-shot models with PQS can fine-tune their parameter using the same way as in the mete-training phase. PQS has a benefit to assist few-shot models adapt to novel tasks.   }
    \label{fig_pqs}
\end{figure}

\begin{figure}[t]
    \centering
    \includegraphics[width=0.76\linewidth]{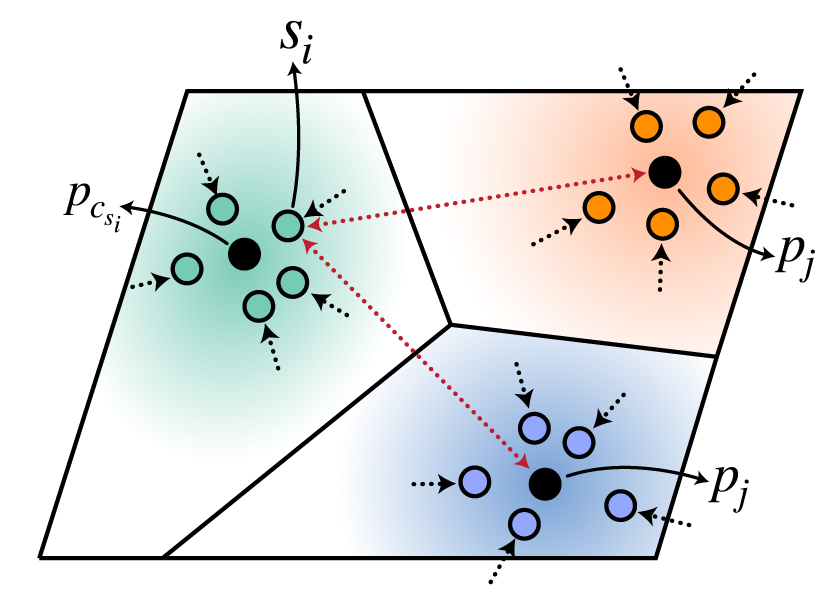}
    \caption{Illustration of PT loss. In PT loss, a sample $s_{i}$ (anchor) is pulled toward its class 
    prototype $p_{c_{s_{i}}}$ (positive) and pushed away from other class prototypes $p_{j}$ (negative).
    Compared to triplet loss \cite{Schroff15Triplet}, we can obtain a more comprehensive margin between each classes, because the sample is operated with prototypes (cf. Section \ref{PT loss}).}
    \label{fig_ptl}
\end{figure}

To formalize the \emph{cross-domain few-shot learning problem}, we follow the
definition in \cite{Guo19ANB}. The \emph{domain} is defined as a joint probability
distribution \(P\) over the input space \(\mathcal{X}\) and label space
\(\mathcal{Y}\). The pair \((x, y)\) represents a sample \(x\) and its corresponding
label \(y\) sampled from \(P\). We denote the marginal distribution of \(\mathcal{X}\)
as \(P_{\mathcal{X}}\). The \emph{base} classes data are sampled from the source
domain (\(\mathcal{X}_{s}, \mathcal{Y}_{s}\)) with joint probability distribution
\(P_{s}\), and \emph{novel} classes data are sampled from the target domain
(\(\mathcal{X}_{t}, \mathcal{Y}_{t}\)) with join probability distribution \(P_{t}\),
and specially \(P_{\mathcal{X}_{s}} \neq P_{\mathcal {X}_{t}}\). A practical example is that models meta-train on a common object dataset (source domain) and meta-test on medical dataset (target doamin). These two domains are not only disjoint but also far away from each other (domain shift). In the meta-testing
phase, models are allowed to be fine-tuned before inferring query samples. During the
fine-tuning , models adapt to the task by training with support set. After that, using
the  accuracy of inferring the category of query samples to evaluate the models
performance. Notably, both the input space and the label space of the source domain
and the target domain are disjoint in the cross-domain few shot learning problem.

\begin{figure*}[t]
    \centering
    \includegraphics[width=0.9\linewidth]{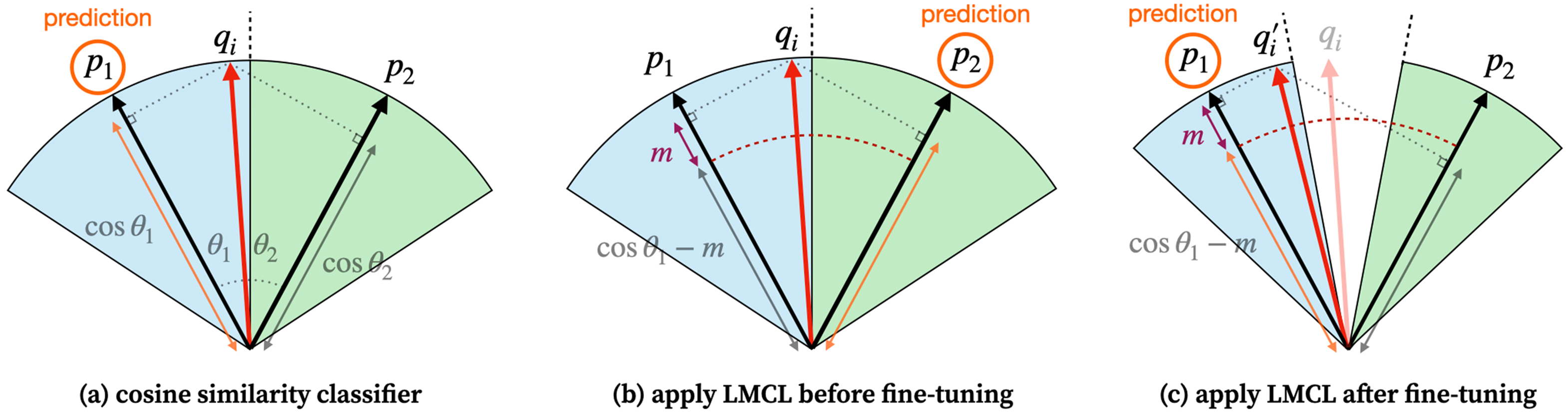}
    \caption{Illustration of LMCL. The classifier needs to categorize a (pseudo) query sample $q_{i}$ to class 1 or
    2. The $p_{1}$ and $p_{2}$ denote the prototype vectors of class 1 and class 2. (a) The cosine
    similarity classifier only compares the magnitude of cosine value and label $q_{i}$ as class 1, 
    because $\cos\theta_{1}$ is bigger than $\cos\theta_{2}$. When two cosine values are close, there is no sufficient margin to categorize.
    (b) LMCL aims to increase the confidence when classifying the sample. LMCL subtracts a margin $m$ on $\cos\theta_{1}$ ($q_{i}$'s class) and then compares with $\cos\theta_{2}$. If the prediction is class 2, then LMCL knows that it doesn't have enough confidence on $q_{i}$. (c) After several fine-tuning iterations, the sample $q_{i}$ is forced to get closer to $p_{1}$ and becomes to $q^{'}$ so as to get larger cosine values . By applying LMCL, the class space of each classes are squeezed, and there is a margin between two classes. The margin is beneficial to classifiers, as they can recognize difference classes easily (cf. Section \ref{section_lmcl}). }
    \label{fig_lmm}
\end{figure*}

\subsection{Overviews} \label{Overviews}
Our motivation is to solve the problem that few-shot models cannot update their parameters by inferring the query set during fine-tuning (query set is reserved for performance evaluation). Thus, we propose the pseudo query set (PQS), By generating
pseudo query set during fine-tuning, few-shot models can be executed the same as in the
meta-training phase and also have the adaptation ability.

In addition, we try fine-tuning the models in a different way, inspired by 
\cite{Chen19Baseline, Guo19ANB}.  In \cite{Chen19Baseline}, Baseline model is trained under a
standard way. Then its trained backbone is extracted and concatenated with a new linear
classifier in the meta-testing phase, using the support set to fine-tune the backbone
and train the classifier. Furthermore, Guo \textit{et al.} \cite{Guo19ANB} concatenates the
backbone with various classifiers, such as cosine similarity or mean-centroid classifier.
We are curious whether the backbone from different models have different performance, and
thus we apply proposed LMM-PQS to backbones from various models. 

In the training phase, we train the Baseline model and meta-train the ProtoNet, respectively.
And LMM-PQS is applied in the meta-testing phase. This phase consists of two stages, fine-tuning
and inference. The process of LMM-PQS is illustrated in Fig \ref{flow_LMM-PQS}. First, the trained
backbone from Baseline or ProtoNet is extracted and used as a feature extractor at both stages. 
During fine-tuning, we generate the pseudo query set (PQS) and fine-tune the backbone with prototypical 
triplet loss (PT loss) and large margin cosine loss (LMCL) several iterations. When inferring, a cosine 
mean-centroid classifier is applied to predict the category of query samples. More precisely, the backbone 
computes the feature embeddings of each (pseudo) query and support sample, which is the $s_{i}$ and $q_{i}$
in Figure \ref{flow_LMM-PQS}, respectively. And the class prototype is the mean value of $s_{i}$ which belong
to the same class. 
Besides, the LMCL and PT loss (see Section \ref{PT loss}) assist the backbone in adapting to the task. At the
inferring stage, the classifier compares the cosine similarities between the embeddings of query samples and
class prototypes to make inference about the category. About the PT loss, we integrate the ``prototype" concept
which is commonly used in many few-shot models and the triplet loss \cite{Schroff15Triplet}, proposing this novel loss function.
On the other hand, the large margin cosine loss is inspired by the similarity between the few-shot learning
problem and open-set problem. The detailed explanation is given in the following sections.

\subsection{Pseudo Query Set (PQS)} \label{section_pqs}
As previously stated, few-shot models need both support sets and query sets to update their parameters.
But the query set is reserved for the performance evaluation in the meta-testing phase, inferring the
query set is prohibited during fine-tuning. To solve the problem that few-shot models can't be fine-tuned due
to the lack of the necessary components, we leverage support sets and a sequence of digital image processing
operations to generate pseudo query sets. Those operations consist of \emph{gamma correction}, \emph{random
erasing with mean RGB values}, \emph{color channel shuffle}, \emph{flip} and \emph{rotation}.
The applying probability of  \emph{gamma correction} and \emph{color channel shuffle} is 0.3, and the probability
of rest operations is 0.5. 

As shown in Figure \ref{fig_pqs}, these operations are randomly applied, and each support image may be used
to generate single or multiple pseudo query image(s) according to the number of support images, detailed
number is demonstrated in the experiment section. With pseudo query sets, few-shot models can update
the parameters during the fine-tuning stage, as they do with normal query sets in the meta-training phase. We can see the benefit when applying PQS to fine-tune few-shot models form Section \ref{exp_pqs} and Table \ref{table:pqst}. 

\subsection{Prototypical Triplet Loss (PT loss)} \label{PT loss}
We propose a loss function which is an assemble of prototypes and triplet loss \cite{Schroff15Triplet},
named prototypical triplet loss. This loss function pulls support samples of each class closer to their
prototype and push them away from prototypes of other classes. Compared to triplet loss, PT loss can 
obtains a more complete margin between each classes. Because the anchor is operated with class 
prototypes, the ability to move toward the center of the category will be stronger than original 
triplet loss.  To be more specific, we first calculate the prototypes for each class within the support
set. Then, each support sample (anchor) and its class prototype (positive) are paired with prototypes 
of all other classes (negative). Subsequently, the original triplet loss is applied. Because there are 
\emph{N} classes in the support set, each sample get \emph{N-1} triplet loss values. Afterwards, we 
loop over the samples and sum up all the values as 
the PT loss, which can be 
formalized as, 
\begin{equation}
PT loss = \sum_{i=1}^{N \times K}\sum_{j=1}^{N} \;(c_{s_{i}} \neq j) \;triplet(s_{i},\;
p_{c_{s_{i}}},\;p_{j})
\end{equation}
where \(i\) is the sample index and \(j\) is the class index, respectively, and \(c_{s_{i}}\) is the class
index of sample \(s_{i}\). The function \(triplet( )\) is the original triplet loss, and \(p_{c_{s_{i}}}\)
is the prototype of the class of \(s_{i}\), \(p_{j}\) is the prototype from other classes. Figure 
\ref{fig_ptl} is the schematic diagram of PT loss. The support sample $s_{i}$ is closer to its class
prototype $p_{c_{si}}$ after applying PT loss and is far away from other prototypes. After fine-tuning the 
backbone with PT loss, we expect that it can recognize different categories more easily. This loss function
belongs to the large margin mechanism introduced in Section \ref{related_lmm}.

\begin{table}[!t] 
\begin{center}

\caption{Setting of hyper-parameters used in meta-testing phase }
\label{t_hps}
\begin{tabular}{ccc}
\toprule
hyper-parameters & usage & values\\
\midrule
$lr$ & learning rate of Adam optimizer & 0.01\\
$N_{ft}$ &\# of fine-tune epoch & 100 \\
$margin$ & \emph{margin} in PT loss & 1.0\\
$s$ & \emph{scale} in LMCL & 30.0 \\
$m$ & \emph{margin} in LMCL & 0.35 \\
\bottomrule
\end{tabular}
\end{center}

\end{table}

\subsection{Large Margin Cosine Loss (LMCL)} \label{section_lmcl}
When solving a problem, models are usually evaluated under closed-set setting or open-set setting. For the
closed-set setting, all testing classes are predefined in the training data, which means the label space of test classes is as same as the label space of training classes.  Hence, it can be regard as a
general classification problem. On the other hand, under the open-set setting, the test classes and the 
training classes are usually disjoint. Thus, models need to learn how to recognize the 
\emph{unseen} category in the testing phase. From this perspective, we argue that the cross-domain few-shot 
learning problem is similar to the open-set problem and treat it as near-open-set problem. Consequently,
those large margin mechanisms should also work in the cross-domain few-shot learning problem.

In open-set problems, many models \cite{Schroff15Triplet, Wang18CosFace, Deng19ArcFace} apply large margin
mechanisms to promote their performance. Motivated by these approaches, we apply large margin cosine loss (LMCL) as one of the losses during the fine-tuning stage, helping the model to recognize
different \emph{novel} categories. In our LMCL, we replace the normalized weight vector with
each class prototype's embedding. As illustrated in Figure \ref{fig_lmm}, when we categorizing $s_{i}$
using cosine value between normalized embedding of  $s_{i}$ and normalized embedding of class prototype
$p_{1}$ or $p_{2}$, we need to have enough confidence (margin). If $s_{i}$ belongs to class $1$ (red class),
and the angle between $s_{i}$ and class prototype $p1$ is $\theta_{1}$ , then LMCL subtracts the cosine value 
$cos\theta_{1}$ with a margin, the subtracted cosine value  still need to be the maximum, or the classification 
is failed. Hence, LMCL can squeeze the class space, because each sample is required to approach their
prototype for a larger cosine value. Thus, the margin between classes is enlarged after applying LMCL.

\section{Experiments} \label{Experiments}
\begin{figure*}[t]
  \centering
    \centering\includegraphics[width=\linewidth]{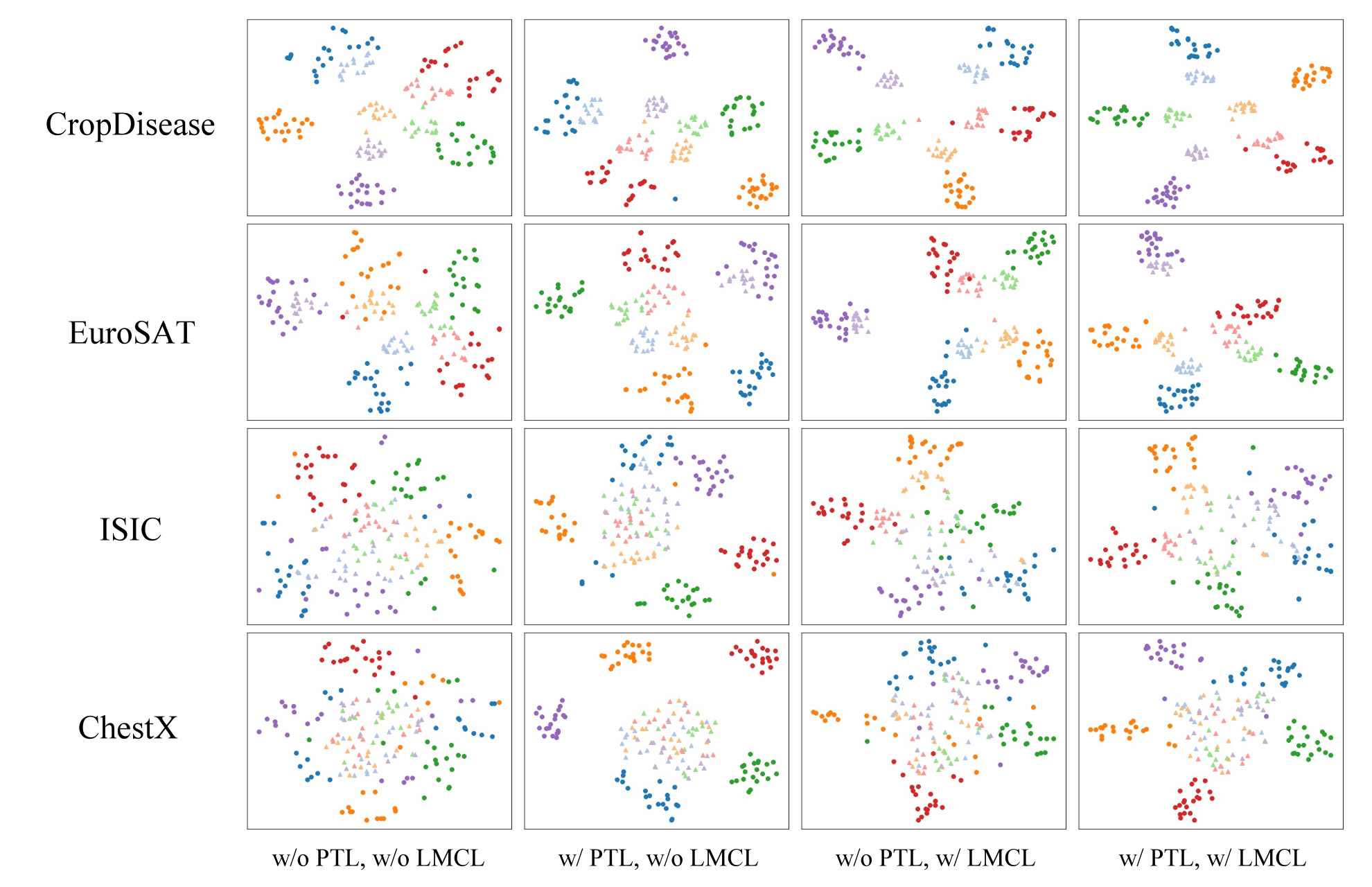}
  \caption{t-SNE results of support set and query set conducted on the 5-way 20-shot setting (four different domains).
  The circles (dark colors) represent support sets and the triangles (light colors) stand for query sets. The five different colors indicate five classes. Adding PT loss and LMCL can aggregate the data from the same classes closer and separate the data from different classes farther (best see in color, cf. Section \ref{section_ablation}). }
  \label{fig: tsne}
\end{figure*}

\begin{table*}
\begin{center}
\caption{ The importance of pseudo query set (PQS). The results of three common few-shot models w/ and
w/o pseudo query set are listed. Evidently, few-shot model fine-tuning with PQS obtain higher
performance. In general, the accuracy of few-shot models with PQS can increase 3$\%$-20$\%$, depending on difference datasets. The reason why the accuracy improvements are significant is that few-shot models with PQS are able to adapt to tasks from novel domains. See Section \ref{exp_pqs} for detailed discussion.}
\label{table:pqst}
\begin{tabular}{ccccccc}
\toprule
\multirow{2}{*}{backbones} & \multirow{2}{*}{ models} & \multirow{2}{*}{PQS} & \multicolumn{4}{c}{5-way 5-shot}\\
&&& CropDiseases & EuroSAT & ISIC &  ChestX\\

\midrule
\multirow{6}{*}{ResNet10}
& \multirow{2}{*}{\small MatchingNet \cite{Vinyals16MN}}
&  & \small 76.46\% \footnotesize (0.71) & \small 70.28\% \footnotesize(0.67) & \small 37.88\% \footnotesize(0.52) & \small 23.34\% \footnotesize(0.39)\\
&& \footnotesize\checkmark & \small \textbf{90.91\% \footnotesize(0.47)} & \small \textbf{81.85\% \footnotesize(0.61)} & \small \textbf{44.20\% \footnotesize(0.58)} & \small \textbf{25.12\% \footnotesize(0.41)}\\
\cline{2-7}
\\[-0.5em]
& \multirow{2}{*}{\small ProtoNet \cite{Snell17PN}}
&  & \small 80.94\% \footnotesize(0.65) & \small 73.75\% \footnotesize(0.73) & \small 43.28\% \footnotesize(0.59) & \small 24.32\% \footnotesize(0.42)\\
&& \footnotesize\checkmark & \small \textbf{91.50\% \footnotesize(0.45)} & \small \textbf{80.88\% \footnotesize(0.66)} & \small \textbf{46.90\% \footnotesize(0.58)} & \small \textbf{25.83\% \footnotesize(0.41)}\\
\cline{2-7}
\\[-0.5em]
& \multirow{2}{*}{\small RelationNet \cite{Sung18RN}} 
& & \small 68.86\% \footnotesize(0.75) & \small 61.52\% \footnotesize(0.63) & \small 36.64\% \footnotesize(0.56) & \small 23.62\% \footnotesize(0.43)\\
&& \footnotesize\checkmark & \small \textbf{90.50\% \footnotesize(0.48)} & \small \textbf{81.18\% \footnotesize(0.63)} & \small \textbf{45.56\% \footnotesize(0.58)} & \small \textbf{25.15\% \footnotesize(0.43)}\\

\bottomrule
\end{tabular}
\end{center}
\end{table*}

\begin{table*}[!t]
\begin{center}
\caption{Ablation study of the proposed LMM-PQS. In most cases, adding both large margin methods will have the highest accuracy. We conduct the experiment started with PQS, because we can't apply fine-tuning in a few-shot style without PQS. See comprehensive discussion in Section \ref{section_ablation}.}
\label{table:ablation}
\begin{tabular}{ccccccccc}
\toprule
\multirow{2}{*}{ backbone } & \multirow{2}{*}{ models } & \multirow{2}{*}{ PQS } &  \multirow{2}{*}{PT loss} &
\multirow{2}{*}{ LMCL } & \multicolumn{4}{c}{ 5-way 20-shot } \\
& & & & & CropDiseases & EuroSAT & ISIC & ChestX\\
\midrule
\multirow{4}{*}{\small ResNet10} & \multirow{4}{*}{\small Baseline} &  \small \checkmark & & & \small 97.48\% \footnotesize(0.24) & \small 92.15\% \footnotesize(0.32) & \small 64.13\% \footnotesize(0.58) & \small 32.74\% \footnotesize(0.47)\\
& &  \small \checkmark & \small \checkmark & & \small 97.50\% \footnotesize(0.22) & \small 92.55\% \footnotesize(0.32) & \small 64.73\% \footnotesize(0.58) & \small \textbf{33.14\% \footnotesize(0.47)}\\
& &  \small \checkmark & & \small \checkmark & \small 97.47\% \footnotesize(0.22) & \small 92.29\% \footnotesize(0.33) & \small 64.71\% \footnotesize(0.56) & \small 32.42\% \footnotesize(0.46)\\
& &  \small \checkmark &  \small \checkmark &  \small \checkmark & \small \textbf{97.60\% \footnotesize(0.23)} & \small \textbf{92.59\% \footnotesize(0.31)} & \small \textbf{64.88\% \footnotesize(0.58)} & \small 32.58\% \footnotesize(0.47)\\
\bottomrule

\end{tabular}
\end{center}
\end{table*}

\begin{table*}[!t]
\begin{center}
\caption{The results of baseline methods and LMM-PQS. We have several observations: (1) LMM-PQS suppress other classifiers in most cases. (2) LMM-PQS with Baseline (ResNet10 backbone) get higher result
on 5-shot and has competitive results on 20-shot or 50-shot when comparing to LMM-PQS with ProtoNet. (3) LMM-PQS with Baseline get better performance with ResNet18 rather than ResNet10. On
the other hand, LMM-PQS with ProtoNet (ResNet18 backbone) reach higher accuracy in first two datasets and has a slightly performance drop in remain datasets. A detailed description is provided in Section \ref{section_comp}.  }

\label{table:resutls}

\begin{tabular}{ccccccccc}
\toprule
\multirow{2}{*}{backbone} & \multirow{2}{*}{models} & \multirow{2}{*}{classifier} & \multicolumn{3}{c}{CropDiseases} & \multicolumn{3}{c}{EuroSAT}\\

& &  & 5-shot & 20-shot & 50shot & 5-shot & 20-shot & 50shot\\
\midrule
\multirow{5}{*}{\small ResNet10} &  \multirow{4}{*}{\small Baseline}
& \footnotesize linear*
& \small 89.25\% \footnotesize(0.51)& \small 95.51\% \footnotesize(0.31)& \small 97.68\% \footnotesize(0.21)
& \small 79.08\% \footnotesize(0.61)& \small 87.64\% \footnotesize(0.47)& \small 91.34\% \footnotesize(0.37)\\
&& \footnotesize linear-T*
& \small 90.64\% \footnotesize(0.54)& \small 95.91\% \footnotesize(0.72)& \small 97.48\% \footnotesize(0.56)
& \small 81.76\% \footnotesize(0.48)& \small 87.97\% \footnotesize(0.42)& \small 92.00\% \footnotesize(0.56)\\
& & \footnotesize mean*
& \small 87.61\% \footnotesize(0.47)& \small 93.87\% \footnotesize(0.68)& \small 94.77\% \footnotesize(0.34)
& \small 82.21\% \footnotesize(0.49)& \small 87.62\% \footnotesize(0.34)& \small 88.24\% \footnotesize(0.29)\\
& & \footnotesize cosine*
& \small 89.15\% \footnotesize(0.51)& \small 93.96\% \footnotesize(0.46)& \small 94.27\% \footnotesize(0.41)
& \small 81.37\% \footnotesize(1.54)& \small 86.83\% \footnotesize(0.43)& \small 88.83\% \footnotesize(0.38)\\
& & \footnotesize LMM-PQS 
& \small \textbf{93.52\% \footnotesize(0.39)}& \small 97.60\% \footnotesize(0.23)& \small 98.24\% \footnotesize(0.17)
& \small \textbf{86.30\% \footnotesize(0.53)}& \small \textbf{92.59\% \footnotesize(0.31)}& \small 94.16\% \footnotesize(0.28)\\
& \small ProtoNet & \footnotesize LMM-PQS
& \small 93.14\% \footnotesize(0.40)& \small \textbf{97.75\% \footnotesize(0.20)}& \small \textbf{98.63\% \footnotesize(0.16)}
& \small 84.24\% \footnotesize(0.57)& \small 92.21\% \footnotesize(0.33)& \small \textbf{94.21\% \footnotesize(0.27)}\\

\hline
\\[-0.5em]
\multirow{2}{*}{\small ResNet18}
& \small Baseline & \footnotesize LMM-PQS
& \small \textbf{94.24\% \footnotesize(0.41)}& \small 97.80\% \footnotesize(0.21)& \small 98.76\% \footnotesize(0.13)
& \small \textbf{86.44\% \footnotesize(0.52)}& \small \textbf{93.17\% \footnotesize(0.32)}& \small 95.29\% \footnotesize(0.29)\\
& \small ProtoNet & \footnotesize LMM-PQS
& \small 93.71\% \footnotesize(0.44)& \small \textbf{97.98\% \footnotesize(0.20)}& \small \textbf{99.08\% \footnotesize(0.12)}
& \small 85.80\% \footnotesize(0.53)& \small 92.68\% \footnotesize(0.33)& \small \textbf{96.01\% \footnotesize(0.32)}\\

\bottomrule
\toprule
\multirow{2}{*}{backbone} & \multirow{2}{*}{models} & \multirow{2}{*}{classifier} & \multicolumn{3}{c}{ISIC} & \multicolumn{3}{c}{ChestX}\\

& &  & 5-shot & 20-shot & 50shot & 5-shot & 20-shot & 50shot\\
\midrule
\multirow{5}{*}{\small ResNet10} &  \multirow{4}{*}{\small Baseline}
& \small linear*
& \small 48.11\% \footnotesize(0.64)& \small 59.31\% \footnotesize(0.48)& \small 66.48\% \footnotesize(0.56)
& \small 25.97\% \footnotesize(0.41)& \small 31.32\% \footnotesize(0.45)& \small 35.49\% \footnotesize(0.45)\\
&& \small linear-T*
& \small 49.68\% \footnotesize(0.36)& \small 61.09\% \footnotesize(0.44)& \small 67.20\% \footnotesize(0.59)
& \small 26.09\% \footnotesize(0.96)& \small 31.01\% \footnotesize(0.59)& \small 36.79\% \footnotesize(0.53)\\
& & \small mean*
& \small 47.16\% \footnotesize(0.54)& \small 56.40\% \footnotesize(0.53)& \small 61.57\% \footnotesize(0.66)
& \small 26.31\% \footnotesize(0.42)& \small 30.41\% \footnotesize(0.46)& \small 34.68\% \footnotesize(0.46)\\
& & \small cosine*
& \small 48.01\% \footnotesize(0.49)& \small 58.13\% \footnotesize(0.48)& \small 62.03\% \footnotesize(0.52)
& \small \textbf{26.95\% \footnotesize(0.44)}& \small 32.07\% \footnotesize(0.55)& \small 34.76\% \footnotesize(0.55)\\
& & \footnotesize LMM-PQS
& \small \textbf{51.88\% \footnotesize(0.60)}& \small \textbf{64.88\% \footnotesize(0.58)}& \small 69.46\% \footnotesize(0.58)
& \small 26.10\% \footnotesize(0.44)& \small 32.58\% \footnotesize(0.47)& \small \textbf{38.22\% \footnotesize(0.52)}\\

& \small ProtoNet & \footnotesize LMM-PQS
& \small 50.57\% \footnotesize(0.63)& \small 63.58\% \footnotesize(0.55)& \small \textbf{69.70\% \footnotesize(0.53)}
& \small 26.07\% \footnotesize(0.43)& \small \textbf{32.72\% \footnotesize(0.50)}& \small 37.91\% \footnotesize(0.48)\\

\hline
\\[-0.5em]
\multirow{2}{*}{\small ResNet18}
& \small Baseline & \footnotesize LMM-PQS
& \small \textbf{52.26\% \footnotesize(0.60)}& \small \textbf{65.84\% \footnotesize(0.58)}& \small \textbf{72.98\% \footnotesize(0.73)}
& \small \textbf{26.54\% \footnotesize(0.44)}& \small \textbf{34.50\% \footnotesize(0.46)}& \small \textbf{37.04\% \footnotesize(0.52)}\\

& \small ProtoNet & \footnotesize LMM-PQS
& \small 50.60\% \footnotesize(0.63)& \small 62.72\% \footnotesize(0.58)& \small 68.21\% \footnotesize(1.11)
& \small 26.44\% \footnotesize(0.44)& \small 32.81\% \footnotesize(0.47)& \small 35.47\% \footnotesize(0.74)\\

\bottomrule
\end{tabular}
\end{center}
\end{table*}

\subsection{Experiment Setting}
We use the benchmark proposed in \cite{Guo19ANB} to evaluate LMM-PQS performance \footnote{We follow agnostic setting as mentioned (not accessing query set during fine-tuning).}.
In this benchmark, models need to be trained or meta-trained on \emph{mini-ImageNet} dataset
and meta-tested on various domains datasets, including \emph{CropDisease}, \emph{EuroSAT}, \emph{ISIC}
and \emph{ChestX} datasets, which contains plant disease images, satellite images, dermoscopic
images of skin lesions and X-ray images, respectively. The sequence of four test datasets are listed
from similar to dissimilar compared to \emph{mini-ImageNet}. For few-shot learning problem, this
benchmark reflect several practice cases in real world since collecting sufficient data from
aforementioned domains are usually difficult and costly. Moreover, this benchmark brings the 
challenge to the few-shot models, since there is a huge domain shift between training and testing
domains. How to adapt the trained model to the test tasks is critical. This domain shift include from
perspective to no perspective, from color images to grayscale images and from natural photo to medical
images.

All rules in the benchmark are followed, we train the Baseline and meta-train the few-shot models,
respectively. Moreover, hyper-parameters used in the training phase are the same in \cite{Guo19ANB},
except the number of tasks for few-shot models is changed from 100 to 300. In the meta-testing phase, models are evaluated on three different settings, including 5-way 5-shot, 5-way 20-shot, and 5-way 50 shot for each dataset.
Besides, same 600 random tasks are sampled from each dataset. Models are evaluated on these tasks, and
the average accuracy with $95\%$ confidence interval are reported. For the hyper-parameters used in the
meta-testing phase, we illustrate all values in Table \ref{t_hps}. All experiments are conducted on a 64-bits Linux machine with Intel I9-9900K CPU and two Nvidia RTX 2080ti GPU cards.

We select Adam optimizer to train and fine-tune the models. Moreover, the backbone architecture is ResNet10
\cite{He16ResNet} in all experiments for fair comparison. Meanwhile, we also provide results with other
backbones. The size of pseudo query set differs in different evaluation setting. For 5-shot, each support
sample generates 4 pseudo query images, and the size of the pseudo query set is 100. For 20-shot, each support
sample produces 2 pseudo query images, and the size of the pseudo query set is 200. Finally, for the 50-shot,
we select 40 support samples from each class and each sample generates 1 pseudo query image, and thus the size 
of the pseudo query set is also 200. In addition, the transductive inference mentioned in \cite{Guo19ANB} are applied in all experiments, which means BatchNorm layers can learn the implicit information from query set when inferring. 

\subsection{Few-shot Models with PQS Results} \label{exp_pqs}
In this section, we discuss the importance of the pseudo query set. The results of three few-shot models are
shown in Table \ref{table:pqst}, including MatchingNet \cite{Vinyals16MN},  ProtoNet (PN) \cite{Snell17PN} and
RelationNet (RN) \cite{Sung18RN}.  The check mark in the \emph{PQS} columns indicates that the model is
fine-tuned with pseudo query set or infer the query set directly. 

Every models with PQS can increase the performance $10\% - 20\%$ in first two datasets. In addition, it is not 
surprised that the performance margin w/ and w/o pseudo query set is relatively small in last two dataset, 
according to larger domain shifts from training dataset. But, it is still obvious that few-shot models 
fine-tuning with pseudo query set gets better performance rather than inferring the task directly. When 
comparing between few-shot models with PQS, ProtoNet get highest accuracy in three datasets. On the other
hand, MatchingNet and RelationNet get competitive results.

\begin{figure}[t]
  \centering
    \centering\includegraphics[width=\linewidth]{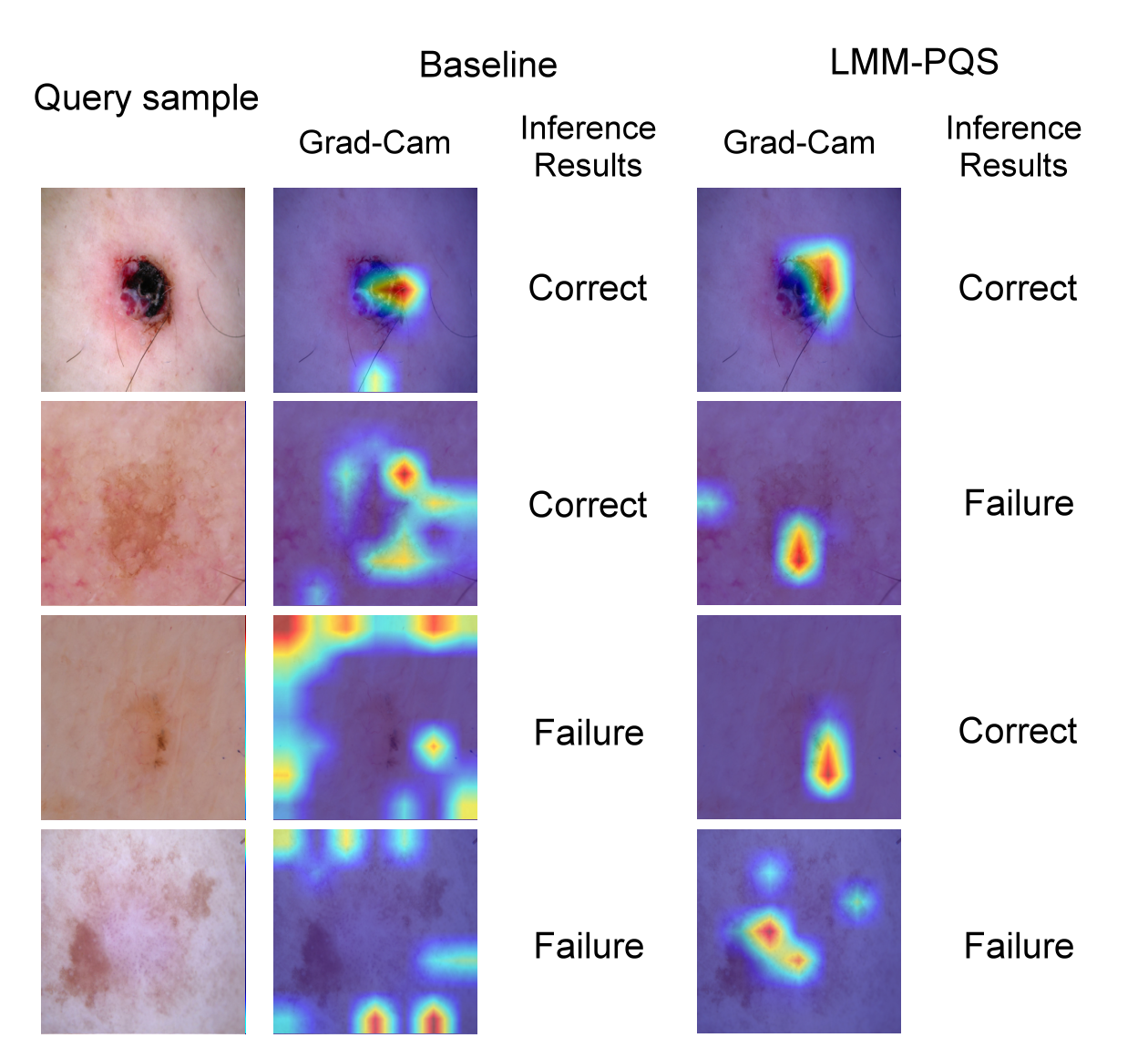}
  \caption{Case discussion. The Grad-Cam \cite{Selvaraju19Grad} results of Baseline and proposed LMM-PQS. It's obvious that LMM-PQS focuses on the important part (red region means higher gradient) in the query image even if the inference result is a failure. The result turns out that training with our LMM-PQS are more robust. On the other hand, the gradient of Baseline model with linear classifier is more dispersed (cf. Section \ref{section_discussion}).  }
  \label{fig_discussion}
\end{figure}

\subsection{Ablation Study} \label{section_ablation}
In this part, we are going to discuss the impact of each components in proposed LMM-PQS. Since that the performance of 5-shot varies a lot than other settings, and the 50-shot setting costs too much time. Hence, we conduct the ablation study on the 20-shot setting. 

As previous stated, we start the ablation study with PQS due to the few-shot style fine-tuning. Table \ref{table:ablation} shows the results. We can see that either adding PT loss or LMCL can slightly increase the performance. Additionally, adding PT loss can improve more than LMCL. In most cases, adding both large margin methods will have the highest accuracy. Furthermore, 

Figure \ref{fig: tsne} shows the visualization result about latent vectors of support set (dark colors) and query set (light colors). It is obvious that when implementing our core techniques, the model aggregates the data from the same classes closer and separates the data from different classes farther, especially in CropDisease and EuroSAT.
More precisely, either implementing PT loss or LMCL, the difference between the intra-class and inter-class distance of support sets increase. 

To sum up, in first three datasets, adding PT loss and LMCL will improve the accuracy, which also reflect on the result of T-SNE. However, in ChestX dataset, although the accuracy slightly improved after adding PT loss, the query set in visualization still cannot be separated well.

\subsection{Comparison with Previous Methods} \label{section_comp}
We compare LMM-PQS with several classifiers evaluated in \cite{Guo19ANB} and illustrate the results in Table \ref{table:resutls}. Moreover, we also measure the performance of different backbone resources (Baseline or ProtoNet) or architectures (ResNet10 or ResNet18). And the result of classifiers with * in the column 
\emph{classifier} is produced in \cite{Guo19ANB}. The linear, linear-T, mean and  cosine term represents linear classifier,
linear classifier with transductive inference, mean-centroid classifier and cosine similarity classifier, respectively. Besides, LMM-PQS contains all components introduced in Section \ref{methodology}.  

First, we observe that LMM-PQS outperforms other classifiers with significant improvement in most cases, except for the
ChestX 5-shot. This result demonstrates LMM-PQS has the powerful ability to assist backbones in adapting to novel tasks.
Then, we compare the performance of backbones from different resources. Overall, LMM-PQS with Baseline gets higher 
performance in most cases on the 5-shot setting, but there is no significant difference in the results on 20-shot or 
50-shot setting.  Moreover, we further investigate the performance from different backbone architectures. For the 
LMM-PQS with Baseline, it is not surprised that ResNet18 obtains higher accuracy against ResNet10 in general. But, the 
result of the LMM-PQS with ProtoNet is interesting. It can get the best performance in first two datasets, but the 
ResNet18's result is even worse than the ResNet10's results in last two datasets. Therefore, in the cross-domain 
few-shot learning, it isn't satisfied that using deeper network architecture will get better performance in common computer vision problems.

\section{Discussions} \label{section_discussion}
In this section, we want to compare which part of the image the Baseline model with linear 
classifier and LMM-PQS focus on. Thus we generate a task from
ISIC dataset and let both models to solve the task. Meanwhile, we apply Grad-Cam \cite{Selvaraju19Grad} to 
visualize the focal point (red region means higher gradient) of the query images. In Figure \ref{fig_discussion}, we demonstrate four different inference result cases of this task. 
For the case which both model can infer correctly, both models can focus on the important region.
For other three cases, the Grad-cam results of Baseline model are messy. Compared to Baseline model, our LMM-PQS can still stick on the foremost area even if the inference is failure. It reveals that LMM-PQS is more robust and has the ability to focus on the critical part of the images.

Why our LMM-PQS still make wrong inference even the model had focused on the important region? We speculate that this is  because samples from the classes in the ISIC dataset are similar to each other in many cases. For example, the initial symptoms of many skin diseases look the same, so even LMM-PQS focus on the critical region, it still has the chance to infer the sample to similar but wrong category.

\section{Conclusion}
In this paper, we tackle the \emph{cross-domain few-shot learning problem} and propose pseudo query set
to solve the problem that few-shot models can't infer query set during fine-tuning. We observe that few-shot
models can get outstanding results after several fine-tuning iterations. According to the results, it shows
that few-shot models still need appropriate fine-tuning when there is a large \emph{domain shift} between the
domains of \emph{base} classes and \emph{novel} classes. In addition, we try fine-tuning the backbones extracted
from the models with large margin mechanisms, including PT loss and LMCL, and surprisingly found that the backbone
performance from Baseline and few-shot models are competitive under same network architecture. We conclude that the
backbone trained in a standard way has robustness. Even if the parameter updating switched to the few-shot style, the
backbone can still adapts to the tasks rapidly. Experiment results illustrate LMM-PQS has a significant performance
improvement compared to the baseline methods. Moreover, how to improve the performance on \emph{ChestX} dataset needs
further investigation.

\ifCLASSOPTIONcaptionsoff
  \newpage
\fi



\bibliographystyle{IEEEtran}
%

\bibliography{main}

%




\end{document}